\documentclass[runningheads]{llncs}

\begin{document}
\title{Conscious Enactive Computation}

\author{Daniel Estrada}
\authorrunning{Estrada}
%
\institute{New Jersey Institute of Technology, Newark NJ 07102 \\
\email{djestrada@gmail.com}}
\maketitle              
\begin{abstract}
This paper looks at recent debates in the enactivist literature on computation and consciousness in order to assess major obstacles to building artificial conscious agents. We consider a proposal from Villalobos and Dewhurst (2018) for enactive computation on the basis of organizational closure. We  attempt to improve the argument by reflecting on the closed paths through state space taken by  finite state automata. This motivates a defense against Clark's recent criticisms of ``extended consciousness'', and perhaps a new perspective on living with machines. 

\end{abstract}
\section{Introduction}
Enactivism challenges the dominant cognitive paradigm in psychology with an account of intentional (purposive) agency that is grounded in the emergent dynamics of biological complexity \cite{di_paolo_enactive_2014,thompson_life_2004,varela_embodied_1991}. Specifically, enactivism holds that biological life is characterized by \textit{adaptive self-constitution}: living systems construct and maintain their own organized structure through their active engagement with a changing world \cite{bechtel_biological_2007,mossio_organisational_2010}. This approach motivates a systematic account of autonomy \cite{barandiaran_autonomy_2017,moreno_biological_2015,ruiz-mirazo_basic_2004,vernon_enaction_2010}, intentional agency \cite{di_paolo_autopoiesis_2005,merritt2015thinking}, subjective consciousness \cite{froese_phenomenology_2010,lutz_neurophenomenology_2003}, and identity in complex dynamical systems \cite{bechtel2012identity,bechtel2017systems}, with the promise of a consistent and unified explanatory framework across the full range of biological processes, from the biomechanics of single-celled organisms to ecologies and societies \cite{froese_enactive_2011,kirchhoff_where_2017,thompson2010mind}.

Despite the emphasis on biological complexity, enactivism has from its inception maintained a robust research program investigating artificial intelligence, artificial life, and robotics (hereafter AI) \cite{agmon_biological_2018,aguilar_past_2014,de_loor_enaction-based_2009,di2017sensorimotor,froese_enactive_2009,suzuki2008enactive}. This research aims to develop models, simulations, and robots that assist in the scientific investigation of biological complexity and adaptive systems. For instance, AI that exhibits some dynamically self-organizing behavior might serve as a useful ``proof of concept'' demonstrating key enactivist principles (see \cite{froese_enactive_2009} for examples). However, while robotics research has already felt a significant impact from the embodied approach \cite{pfeifer2005interacting,pfeifer2005new}, enactivist AI is often advanced against a backdrop of criticism directed at ``merely'' computational or representational explanations \cite{hutto_radicalizing_2012,hutto_evolving_2017}. As a founder of enactivism Francisco Varela put it, ``This fundamental paradigm of the digital computer program will not do for biology, nor for AI.'' \cite{varela_embodied_1991}

A recent set of papers from Villalobos, Dewhurst, Ward and colleagues (hereafter Villalobos)  \cite{dewhurst_enactive_2017,villalobos_computationalism_2016,villalobos_why_2017,villalobos_enactive_2018} address these historical tensions between enactivism and computation. Villalobos argues that the enactivists are mistaken to treat computers as mere symbolic processors of abstract representations. Drawing on a mechanist account of computation, Villalobos suggests an interpretation of the classical Turing machine which they claim would meet enactivist conditions for \textit{self-determination}. If so, it would suggest that embodied agency could be given a computational rather than biological basis without sacrificing enactivist commitments to the dynamical interactions between agent and world. This argument strikes at the foundations of the enactivist program, and threatens to overturn 20+ years of enactivist thought on AI and computation.

The central concern of this paper is to assess the proposal for enactive computation put forward by Villalobos. Their argument turns on the enactivist interpretation of self-determination in terms of \textit{organizational closure}. While we think Villalobos' examples fail to meet strong enactivist conditions on closure, we suggest they can be improved through explicit consideration of the structure of the finite state automata (\textbf{FSM}) that controls a classic Turing machine. This highlights an important form of closure that is, we argue, more fundamental than organizational closure: namely, the closed path through state space taken by the FSM. We claim that computation is fundamentally concerned with the structure of paths through state space, and that all living organisms can be characterized by such paths. This result suggests computation as the fundamental basis from which the enactivist program must emerge. We then consider the implications of this argument for a particular strand of criticism raised by Clark \cite{clark_spreading_2009,clark_whatever_2013} against enactivist proposals for ``extended consciousness'' \cite{noe2004there}. We conclude with general thoughts on the implications these arguments have for living with machines.\par

\section{Organizational closure and Turing's machine}
Organizational closure serves as the basis for the enactivist approach to autonomous intentional (purposive) behavior, and names the sense in which biological organisms are \textit{self-determined}  \cite{barandiaran_autonomy_2017,bechtel_biological_2007,varela1974autopoiesis}. A system is \textit{organized} when its constitutive components are arranged into a network of functionally interdependent processes and constraints \cite{levy2013abstraction}. An organization is \textit{closed} when the operation of its constitutive components are themselves sufficient for the adaptive construction and generation of its organized state \cite{mossio_organisational_2010}. Enactivists argue that organizational closure provides an \textit{intrinsic} basis for identifying organisms and their boundaries as unified wholes. Furthermore, enactivists emphasize that organisms are precariously situated within a dynamic world to which they must continually adapt in order to maintain an organized state. This \textit{precariousness} creates conditions that demand coordinated action from the organism as a whole in order to maintain its organized state \cite{burge2009primitive}. This gives rise to what enactivists call \textit{adaptive sense-making}, which serves as the basis for investigations into consciousness and phenomenology \cite{froese_phenomenology_2010,lutz_neurophenomenology_2003,thompson_life_2004}. 

Beyond its central role in the enactivist theory of autonomous agency, organizational closure also figures in enactivist criticisms of classical computation\footnote{Enactivists are not universally hostile to computation. Importantly, Mossio et al \cite{mossio2009computable} render an organizationally closed system in the $\lambda$-calculus, and argue that ``there are no conceptual or principled problems in realizing a computer simulation or model of closure.'' Such arguments have resulted in a split between radical anti-computationalists \cite{hutto_radicalizing_2012} and more traditional versions of enactivism. See \cite{cardenas2010closure,ward_introduction:_2017}.}. Enactivists contrast the closed structure of biological organisms with the \textit{open} or \textit{linear} structure of traditional computing machines \cite{froese_enactive_2009}. On this view, computers operate through a sequence of formal operations that transforms symbolic ``input'' into symbolic ``output''. Enactvists claim at least two important differences between computation and the adaptive self-constitution of biological organisms. First, computers perform stepwise formal operations on symbolic input, rather than performing dynamic mechanical operations within a changing world. Second, computers don't ``build themselves'' in the sense relevant for adaptive self-constitution, which requires organizational closure. Put simply, computers aren't self-determined wholes with a world of their own, and so cannot serve as the intrinsic subject of an experience. Instead, computers are artifacts created through external processes of human design and manufacturing. Such considerations lead Froese and Ziemke \cite{froese_enactive_2009} to distinguish the \textit{behavioral autonomy} characteristic of certain kinds of self-controlled machines (say, a dishwasher on a timer), from the \textit{constitutive autonomy} characteristic of living biological systems.

Villalobos' argument for enactive computation in \cite{villalobos_enactive_2018} is designed to show that a Turing machine can meet the conditions for self-determination as described by Maturana (1988) \cite{maturana1988ontology}. Here, self-determination is identified with  \textit{functional closure}. A system has functional closure when its organizational structure forms closed feedback loops. As an example, Villalobos offers a thermostat regulating the temperature of a house. The behavior of the thermostat-house system is characterized by a feedback loop between these two components which has a circular structure and satisfies functional closure. Of course, while the thermostat-house system ``controls itself'' with respect to temperature, it is not adaptively self-constituting in any deeper sense; thermostats and houses don't build themselves with their parts alone. Thus, functional closure is not sufficient for organizational closure of the sort required for constitutive autonomy. Nevertheless, Villalobos argues this control structure does not connect inputs to outputs through a linear sequence of symbolic processes, and so is not ``open''. It is, they argue, closed and minimally self-determining in a sense relevant for enactivist theory.

Villalobos then applies this feedback loop model to the classic Turing machine. Turing \cite{turing1937computable} proposed a computing machine with three components: a tape with discrete cells; a read-write head that operates on the tape; and a program which controls the operation of the head. On the enactivist interpretation, the tape serves input to the machine and records output from the machine, and the machine (the head and program) performs formal operations that convert the former to the latter as a linear process. Against this view Villalobos offer an alternative, inspired by Wells \cite{wells1998turing} and Piccinini  \cite{piccinini2007computing,piccinini2015physical}, that interprets the Turing machine in terms of looping interactions between the machine and the tape. This forms a functionally closed loop, much like the thermostat-house system, which implies self-determination in the sense that the computer's state is determined by the interactions between the machine and the tape. In an analog computer these constraints might appear as features of the physical mechanisms of the device, thereby eliminating any symbolic aspect of the computation. Thus, Villalobos argues, even a classical Turing machine can be understood as purely mechanical and functionally closed, and so evades the enactivist criticism of computation. While this argument doesn't entail that computers are conscious living creatures of equivalent complexity to biological organisms, it does confront a major hurdle within the enactivist literature to treating computing machines as genuinely purposive agents with a world of their own. 

Does Villalobos' argument succeed? Briefly, no: functional closure alone is not sufficient for adaptive self-constitution of the sort relevant for intentional agency or adaptive sense-making. Villalobos' `enactive' Turing machine is merely behaviorally and not constitutively autonomous. While Maturana's account is influential, recent work has developed more rigorous constraints on organizational closure. For instance, Mossio et al. \cite{montevil_biological_2015,mossio_organisational_2010} present a model of closure which requires that constitutive constraints operate across multiple scales or levels of organization to achieve closure. While the thermostat-house system is functionally closed, we might say that closure occurs at a single scale, namely the feedback loop that controls temperature. At other scales, for instance the internal structure of the thermostat mechanism, the system is not closed or self-determining. Similarly, Turing's machine appears to be functionally closed only at the level of operations of the head on the tape and nowhere else. Biological systems, on the other hand, are in some sense self-determining \textit{all the way through}---or at least they are self-organized across a range of scales from inter-cellular biochemistry through geopolitics that covers the breadth of our experiences of a meaningful world as human agents. A Turing machine might be functionally closed, but it covers nothing close to the same range of interactivity.

How many levels of organizational constraints are required to distinguish between behavioral and constitutive autonomy? Mossio's model suggests at least two. If so, Villalobos' argument might be improved by describing a Turing machine with two layers of self-determining organizational constraints rather than one. In the next section, I will discuss how the classic Turing machine already captures organizational closure across two layers of constraint.

\section{Closed paths through state space}
If we suspend the anti-representational commitments of enactivism for a moment, there's an important feature of Turing's machine which is not explicitly addressed in these arguments: the structure of the program which controls the read-write head. In Turing's model, the program takes the form of a finite state machine (\textbf{FSM}). FSMs are abstract automata that are characterized by a finite number of discrete states, and a set of rules that describe the operations performed in each state, and the conditions for transitioning between states, depending on what is read from the tape. These rules can be represented as a state transition table, which can be realized\footnote{For historical reasons originating with Putnam \cite{hilary1988representation}, it is often taken for granted that a definition of computation in terms of finite state automata cannot distinguish between different \textit{realizations} of a computer, and so cannot in principle provide an explanation for cognitive behavior. Piccinini \cite{piccinini2007computing} cites this as an explicit motivation for developing his mechanistic account of computation. There are good reasons for thinking that Putnam's concerns are overstated  \cite{chalmers1996does,joslin2006real}, but this issue is beyond the scope of this paper. Thanks to Jon Lawhead for pointing this out.} in a physical machine in a number of ways. The physical Turing machine is `programmed' insofar as it realizes the abstract state transition structure of the FSM. 

The abstract nature of the FSM should not worry enactivists \cite{levy2013abstraction}. An FSM can in principle be realized by simple physical mechanisms; there's nothing inherently ``symbolic'' about the FSM. The FSM does not directly concern the relationship between a computer and its environment; the FSM is not (necessarily) used by a computer to represent the world. The FSM is just an abstract model of the states a machine can be in, and the conditions for transitioning between these states. Enactivist literature is often directly preoccupied with systems being in certain states, like the equilibrium state (homeostasis), and with the activities organisms must perform to maintain these states \cite{kauffman2000investigations,ruiz-mirazo_basic_2004}. To this extent, enactivist theory depends on state abstractions of the same sort used to describe the FSM. Describing the autonomy of an organism in terms of ``organizational closure'' is already to appeal to an abstract control structure, so there should be no principled objections from enactivists to discussing the equally abstract structure of the FSM. 

While the FSM can be represented as a transition table, it is also customary to represent an FSM with a \textit{state space diagram} with states represented as circles, and arrows between circles representing the transitions between states. A state space diagram has a \textit{closed path} (or \textit{loop}) if some sequence of operations will return the system to a previous state. For instance, suppose I take water at room temperature, freeze it to ice, then let it thaw back to room temperature. The water changed state, then changed back; we can represent this as a short path through the state space of water that loops back to where it began. Homeostasis is an interesting state for biological organisms precisely because they maintain the state as a \textit{fixed point attractor}, returning to equilibrium after minor disturbances. This is another way of saying that homeostasis is characterized by a closed path in state space  (\textbf{CPSS}). 

With these considerations in mind, we propose that CPSSs, and paths in state space generally, are of fundamental relevance to enactivist models of self-determination. Moreover, CPSSs put computers and organisms on equal ontological footing. Recall the theoretical motivation for appealing to organizational closure to explain autonomy: it provides an intrinsic basis for individuating a system as a unified whole, and so serves as a basis for adaptive sense-making. We claim that a CPSS accomplishes the same theoretical task: organisms can be identified intrinsically as the collection of processes and constraints that walk a CPSS. This definition is intrinsic in the same sense as organizational closure: whether a path counts as ``closed'' is set by the constitution of the system itself. More strongly, we claim that any organizationally closed system can be characterized by a collections of CPSSs with a fixed attractor at the constitutive organized state. This suggests that CPSSs are theoretically a more fundamental form of closure than organizational closure. Indeed, the important sense of `closure' captured by the enactivists has less to do with daisy-chained functions looping on themselves, and more to do with the CPSSs those functional relationships enable. Strictly speaking, neither functional nor organizational closure is necessary for walking a CPSS.

Not every Turing machine will walk a CPSS, but it is exceedingly common for them to do so\footnote{The question of deciding in general whether a path in state space will close is formally equivalent to the halting problem, and so is not computable. See  \cite{luz_cardenas_closure_2010}.}. We can think of the CPSSs which characterize a Turing machine's program as another scale of closure, one which directly controls the looping interactions between head and tape. With two scales of closed loops, this would appear to meet Mossio's stronger constraints on closure, and thus we have shown the classical Turing machine might already constitute an adaptively self-constituting system on enactivist grounds. Or, perhaps more realistically, the depth of closure matters a lot less than what states those functional relationships (closed, shallow, or otherwise) make available for the organism as it walks paths in state space.

\section{Extended consciousness}
To appreciate how CPSSs can be useful to enactivism, consider a recent debate on the bounds of consciousness. Despite his strong influence on enactivism, Clark has pushed back against attempts to locate the processes constitutive of conscious experience in the world \cite{clark_spreading_2009}. Clark argues there is no good reason to do so; the activity constitutive of a conscious experience occurs immediately within patterns of neural firings. Clark advocates for an explanatory approach called ''predictive coding'' which uses ``a hierarchical generative model that aims to minimize prediction error within a bidirectional cascade of cortical processing'' \cite{clark_whatever_2013}. Clark argues that the model works by rapidly updating on the basis of new information. This leaves little bandwidth for external changes to impact the updating model beyond sensory input; the dominant influence on a neuron is simply the activity of other neurons. Thus, Clark argues, it is unlikely that external processes play a constitutive role in conscious experience. 

Ward \cite{ward_enjoying_2012} offers a response to Clark on behalf of enactivists that appeals to multiple layers of interactions between the agent and world. Clark's mistake, on this view, is to localize consciousness to any single process in the organized hierarchy. The appeal to multiple layers should by now be a familiar enactivist move, one Clark rejects as superfluous in this case \cite{clark2012dreaming}. Whatever world-involving processes enactivists believe are important, Clark claims he can account for them with predictive coding. So consciousness appears stuck in the head.  

Clark's alternative doesn't appeal to enactivists because the world-involving aspects of predictive coding appear linear and open, like a computer, rather than closed like an organism. This isn't an accurate perception; the cascade of neural activity develops with looping feedback until the neurons reach stability, so there \textit{are} functionally closed processes; those processes just aren't extended and world-involving. They only involve neurons and their cortical support. Enactivists are attracted to externalism because they view consciousness as inherently world-involving and organizationally closed. Just as with Villalobos' computer, enactivists are hoping to find closure in the organizational structure of the embodied conscious state. Since closure is an indicator of unification and wholeness, enactivists expect neural activity and world-involving processes to demonstrate functional interdependencies. Clark's argument that the neural activity is not functionally dependent on external processes is therefore fatal to the view.

Perhaps CPSSs can help resolve this conflict amicably? If we think about closure in terms of CPSSs we can recover the looping interactions that are inherently world-involving and \textit{closed in state space}, while conceding to Clark that the neural activity is sufficiently explanatory of the functional interactions that give rise to the conscious state. In state space we are no longer confined to a single closed loop spanning organizational levels. Instead, our dynamical activity across different scales will form many different kinds of closed paths in different state spaces. Some of these CPSSs will be characterized by inherently world-involving \textit{states}, and so will recover an enactivist sense of closure compatible with predictive coding. 

Consider, for instance, that it is easier to maintain your balance with your eyes open than closed. Here we have two cortical cascades: one producing visual experiences, and one producing motor activity to maintain balance. These two systems reinforce each other. Maintaining balance is a precarious state that inherently involves the configuration of the body as a massive physical object with specific dimensions. Thus, the configuration of my body is a fundamental factor in whether I am in a balanced state. The balanced state is a fixed attractor for certain CPSSs that characterize my attempts to stay balanced. This brings in looping, inherently world-involving processes into an explanation of my behavior as an agent without committing to implausible functional interdependencies between neurons and world. The important dependencies for closure, and ultimately for autonomy, identity, and consciousness, are found in state space. 

\section{Conclusion}
We don't view CPSSs as a threat to enactivism's positive theory of autonomy or adaptive sense-making. Instead, we see it correcting the over-emphasized anti-computationalism that has historically motivated the view. We think enough speaks in favor of the enactive approach that it needn't appeal to a questionable and increasing problematic ontological distinction between computing machines and biological life. Insofar as Villalobos' argument also serves these goals, this paper is meant to push harder in the same direction.

\bibliographystyle{splncs04}
\bibliography{cce}
\end{document}